\documentclass{article}

\usepackage{microtype}
\usepackage{graphicx}
\usepackage{subfigure}
\usepackage{booktabs} 
\usepackage{amsmath, amssymb, amsfonts, amsthm}
\usepackage{amssymb}
\usepackage{verbatim}
\usepackage[utf8]{inputenc}
\usepackage{pgfplots}
\pgfplotsset{compat=newest}
\usepgfplotslibrary{groupplots}
\usepgfplotslibrary{dateplot}

\usepackage{amsmath,amsfonts,bm}

\def\eqref#1{equation~\ref{#1}}

\def\1{\bm{1}}

\def\vb{{\bm{b}}}

\def\ve{{\bm{e}}}

\def\vg{{\bm{g}}}
\def\vh{{\bm{h}}}

\def\vm{{\bm{m}}}

\def\vo{{\bm{o}}}

\def\vu{{\bm{u}}}

\def\vx{{\bm{x}}}

\def\mA{{\bm{A}}}
\def\mB{{\bm{B}}}
\def\mC{{\bm{C}}}
\def\mD{{\bm{D}}}

\def\mH{{\bm{H}}}
\def\mI{{\bm{I}}}

\def\mU{{\bm{U}}}

\def\mW{{\bm{W}}}

\DeclareMathAlphabet{\mathsfit}{\encodingdefault}{\sfdefault}{m}{sl}
\SetMathAlphabet{\mathsfit}{bold}{\encodingdefault}{\sfdefault}{bx}{n}

\newcommand{\R}{\mathbb{R}}

\usepackage{hyperref}

\usepackage[accepted]{icml2021}

\begin{document}

\twocolumn[
\icmltitle{Parallelizing Legendre Memory Unit Training}



\icmlsetsymbol{equal}{*}
\begin{icmlauthorlist}
\icmlauthor{Narsimha Chilkuri}{uw}
\icmlauthor{Chris Eliasmith}{uw,abr}
\end{icmlauthorlist}

\icmlaffiliation{uw}{Center for Theoretical Neuroscience, University of Waterloo}
\icmlaffiliation{abr}{Applied Brain Research}

\icmlcorrespondingauthor{Narsimha Chilkuri}{narasimha.chilkuri@gmail.com}

\icmlkeywords{Machine Learning, ICML}

\vskip 0.3in
]



\printAffiliationsAndNotice{}  


\begin{abstract}
Recently, a new  recurrent neural network (RNN) named the Legendre Memory Unit (LMU) was proposed and shown to achieve state-of-the-art performance on several benchmark datasets. Here we leverage the linear time-invariant (LTI) memory component of the LMU to construct a simplified variant that can be parallelized during training (and yet executed as an RNN during inference),  thus overcoming a well known limitation of training RNNs on GPUs. We show that this reformulation that aids parallelizing, which can be applied generally to any deep network whose recurrent components are linear, makes training up to 200 times faster. Second, to validate its utility, we compare its performance against the original LMU and a variety of published LSTM and transformer networks on seven benchmarks, ranging from psMNIST to sentiment analysis to machine translation. We demonstrate that our models exhibit superior performance on all datasets, often using fewer parameters.  For instance, our LMU sets a new state-of-the-art result on psMNIST, and uses half the parameters while outperforming DistilBERT and LSTM models on IMDB sentiment analysis. 
\end{abstract}

\section{Introduction}
\label{introduction}


LSTMs \cite{hochreiter1997long}, the most popular class of recurrent neural networks (RNNs), process information in an inherently sequential manner. This prevents parallelization within training examples, and thus, they cannot fully leverage today's commodity GPU hardware. This sequential nature of RNNs is one of the critical reasons for the shift towards purely self-attention based architectures such as the transformer and its derivatives \cite{vaswani2017attention, devlin2018bert, brown2020language}, especially in the domain of Natural Language Processing (NLP). Parallelization makes training such networks far more efficient on GPUs, and thus allows them to be applied to enormous datasets -- for example, the Colossal Clean Crawled Corpus (C4) which comprises 750GB of English text \citep{raffel2019exploring}. Hence, such models, via unsupervised pre-training, make it possible for us to exploit resources such as the internet,\footnote{https://commoncrawl.org} which produces 20TB of text data each month. A feat such as this, from the training perspective, would be unimaginable using LSTM or other RNN based models.

In this paper, we exploit the idea that linear recurrence relations can be {\it solved}. More specifically, a linear time-invariant (LTI) system's state update equation can be written in a non-sequential fashion \cite{aastrom2010feedback}. This allows for the computation of the hidden state of an LTI system to be done in parallel, thus overcoming the fundamental constraint that traditional RNNs suffer from. Despite this desirable property, to our knowledge, such systems have remained underexplored. This perhaps has to do with the general belief that nonlinear dynamics are critical for solving complicated problems, especially AI level tasks. Here, we challenge that notion by focussing our attention on a specific LTI system, the memory component of the recently proposed Legendre Memory Unit \citep{voelker2019legendre}, and demonstrating superior performance of our model on a variety of tasks, including predicting a nonlinear, chaotic dynamical system.

The Legendre Memory Unit (LMU) is an RNN cell that is constructed by coupling an LTI system to a nonlinear one. The LTI system, known as the Delay Network \citep{voelker2018improving}, is the core component of the LMU that projects a sliding window of length $\theta$ of the input signal onto the Legendre basis -- hence the name Legendre Memory Unit. This construction, which employs the Delay Network (DN) to act as the memory and a nonlinear recurrent layer to compute arbitrary functions across time, has been shown to outperform LSTMs and other RNNs on various tasks. Of special interest to us is the ability of the LMU to handle temporal dependencies across 100,000 time-steps, which is orders of magnitude better than the LSTM.

In the following pages, we start by simplifying the LMU such that recurrence exists only in the linear system. After showing that the training of this simplified variant can be parallelized, we turn to the question of its effectiveness. We do so by first considering the two synthetic tasks that the original LMU was validated on: psMNIST and Mackey-Glass. We show that our variant  exceeds the original LMU's performance on both of these tasks, while also establishing a new state-of-the-art result for RNNs on psMNIST. We also test our models against LSTM models of comparable sizes on several NLP tasks. We look at sentiment classification (IMDB), semantic similarity (QQP), and natural language inference (SNLI), and show that our models achieve better performance while using significantly fewer parameters (up to 650x). We then briefly investigate the transfer learning abilities of our model by training a language model on the (unlabelled) Amazon Reviews dataset and using that to improve the IMDB score. Here, we show that it outperforms DistilBert, a transfomer based model, while using 50\% fewer parameters. We conclude our NLP experiments by demonstrating superior performance on language modelling (text8) and machine translation (IWSLT'15 En-Vi). Finally, we show that these architectural modifications result in training times that are up to 200 times faster, relative to the original LMU.

\section{Related Work}

There have been recent studies focused on the application of LMU to new problems and on the modification of the original architecture. For example, \citet{blouw2020hardware} used an LMU based architecture to achieve SotA results on the problem of efficient key-word spotting, and \citet{gu2020hippo} recently proposed a generalization of the original architecture, which they use to beat the SotA result set by the LMU on psMNIST. 

Our model, however, is more directly inspired by the success of self-attention. Self-attention based architectures have come to replace RNN based approaches for problems such as language modelling, machine translation, and a slew of other NLP tasks \citep{radford2018improving, raffel2019exploring}. Three properties that make self-attention desirable over RNNs are: (1) it is better at handling the challenging problem of long-range dependencies; (2) it is purely feedforward; and (3) when the sequence length is smaller than the dimension of representation, which is not uncommon in NLP applications, self-attention is computationally cheaper than an RNN.

Of these three desirable properties, our model inherits the first one from the original LMU, and satisfies properties (2) and (3) by construction. Additionally, the capability to run our model in a recurrent manner during inference can be advantageous in situations where there are memory constraints or where low-latency is critical. 

It should also be noted that prior works such as \citet{balduzzi2016strongly} and \citet{martin2017parallelizing}  have explored the use of linear RNNs for sequence modelling. Our work differs from theirs in a few ways. First, we do not learn the recurrent connections, but instead make use of an LTI system, the DN, that optimally implements the delay operation; the use of frozen connections also leads to a more efficient parallelization scheme.  Second, while previous works restrict the recurrent relations to be elementwise (recurrent weight matrix is diagonal), we use full matrices in the linear system. Finally, we test our model on a wide range of tasks, well beyond the synthetic datasets \citep{martin2017parallelizing} and language modelling \citep{balduzzi2016strongly} considered previously.

\section{Background and LMU Variants}

In this section we start by introducing the delay problem and show how a delay is optimally realized by the DN, an LTI system. We then introduce the LMU which employs a single-input DN coupled to a nonlinear dynamical system to process sequential data. Finally, we introduce our model, which is obtained by simplifying the original architecture. We show how this simplification allows for training to be parallelized (often reducing the computation complexity), while also retaining the ability to handle long range dependencies of the original formulation.

\subsection{Delay Network}
\paragraph{Ideal Delay} A system is said be an ideal delay system if it takes in an input, $u(t)$, and outputs a function, $y(t)$, which is the delayed version of the input. Mathematically, this can be described in the following manner:
\begin{equation}
    y(t) = \mathcal{D}[u(t)] = \begin{cases} 0 \quad t < \theta\\ u(t-\theta) \quad t \geq \theta \end{cases},
\end{equation}
where $D$ is the ideal delay operator and $\theta \in \mathbb{R}$ is the length of the delay. There are two things of note here: (1) the ideal delay system is linear, i.e., for any two functions, $f(t)$ and $g(t)$, and any $a, b \in \mathbb{R}$, it respects the following equation:
\begin{equation}
    \mathcal{D}[af(t) + bg(t)] = a\mathcal{D}[f(t)] + b\mathcal{D}[g(t)];
\end{equation}
and (2) although this problem looks deceptively simple, in fact it takes a system with infinite memory to take in a continuous stream of input (with unspecified frequency content), store it for $\theta$ seconds and then reproduce the entire input without error.

These two considerations tell us that the optimal system that implements delay must be linear and that even the most optimal physical implementation can only approximately realize the delay.  

\paragraph{Approximating Delay} If we are interested in constructing a dynamical system that implements delay, thanks to the linearity constraint, we can narrow our search space from the general system of ODEs of the form:
\begin{align}
    \dot{\vm} = \textbf{f}(\vm, u),\\
    y = \textbf{g}(\vm, u),
\end{align}
to just finding the four matrices $(\mA, \mB, \mC, \mD)$ that define an LTI system:
\begin{align}
    \dot{\vm} = \mA \vm + \mB u,\\
    y = \mC \vm + \mD u.
\end{align}
Following the derivation of the Delay Network in \citet{voelker2018improving}, we start by considering the transfer function of the delay system, which for a SISO system is defined as:
\begin{equation}
    G(s) = \frac{y(s)}{u(s)} = e^{-\theta s},
\end{equation}
where $y(s)$ and $u(s)$ are found by taking the Laplace transform of the input and output functions in time. As expected, this defines an infinite dimensional transfer function, capturing the intuitive difficulty of constructing a continuous delay.

The transfer function can be converted to a finite, causal state space realization \textit{if and only if} it can be written as a proper\footnote{A ratio $\frac{a(s)}{b(s)}$ is said be proper if the order of the numerator does not exceed the order of the denominator.} ratio of finite dimensional polynomials in $s$ \cite{brogan1991modern}. $G(s)$, however, is irrational, i.e., it cannot be written as a proper, finite dimensional ratio.  Therefore,  making an approximation is necessary.

We can achieve an optimal convergence rate (in the least-square error sense) for rational approximants by means of Padé approximants \cite{partington2004some}. Choosing the order of the numerator to be one less than the order of the denominator and accounting for numerical issues in the state-space realization (see \citet{voelker2018improving} for details), gives the following canonical realization:
\begin{align}
    & A_{i,j} = \frac{(2i + 1)}{\theta}\begin{cases} -1 \qquad i < j\\ (-1)^{i-j+1} \quad i \geq j \end{cases}, \label{A_cont}\\
    & B_{i} = \frac{(2i+1)(-1)^{i}}{\theta}, \label{B_cont}\\
    & C_{i} = (-1)^{i} \sum_{l=0}^{i} {i \choose l} {i + l \choose j}(-1)^{l},\label{C matrix}\\
    &D = 0, \quad i, j \in [0, d-1],
\end{align}
where we refer to $d$ as the order of the system. The LTI system $(\mA, \mB, \mC, \mD)$ is what is known as a Delay Network (DN). The order of the system, $d$, and the delay length, $\theta$ are the main hyperparameters to choose when using a DN. Higher order systems require more resources, but provide a more accurate emulation of the ideal delay. Because we have used Padé approximants, each order is optimal for that dimension of state vector $\vm$.

\paragraph{Legendre Polynomials} Suppose we construct a system using the $(\mA, \mB, \mC, \mD)$ matrices define above, and provide it with an input signal, $u(t)$. Given the state $\vm_t$, we can use $\mC$ to decode $u(t - \theta)$ to a degree of accuracy determined by the order of our system, i.e.,
\begin{equation}
    u(t-\theta) \approx \mC^{T} \vm_t.
\end{equation}
Intuitively, given $\vm_t$, it seems possible to decode not only $u(t-\theta)$ but also $u(t-\theta') \,\, \forall  \,\, 0 \leq \theta' \leq \theta$.  This can be done using a slightly modified $\mC$ for a given $\theta'$:
\begin{equation}
    u(t-\theta') \approx \mC (\theta')^{T} \vm_t,
\end{equation}
where 
\begin{equation}
    C_{i}(\theta') = (-1)^{i} \sum_{l=0}^{i} {i \choose l} {i + l \choose j}\left(-\frac{\theta'}{\theta}\right)^{l}, 0 \leq \theta' \leq \theta, \label{shifted legendre}
\end{equation}
and $\mC(\theta' = \theta)$ corresponds to the $\mC$ defined in equation (\ref{C matrix}). Interestingly, the functions in (\ref{shifted legendre}) turn out to be the shifted Legendre polynomials. 


\subsection{Legendre Memory Unit}\label{LMU section}
The LMU is obtained by coupling a single-input delay network to a nonlinear dynamical system. The DN orthogonalizes the input signal across a sliding window of length $\theta$, whereas the nonlinear system relies on this memory to compute arbitrary functions across time. The state update equations that define the LMU are given below: 
\begin{align}
    \displaystyle
    & u_{t} = \ve_{x}^{T} \vx_{t} + \ve_{h}^{T}\vh_{t-1} + \ve_{m}^{T}\vm_{t-1},\label{input} \\
    &\vm_{t} = \Bar{\mA} \vm_{t-1} + \Bar{\mB} u_{t},\label{DN}\\
    &\vh_{t} = f(\mW_{x} \vx_{t} + \mW_{h} \vh_{t-1} + \mW_{m} \vm_{t}), \label{nonlinear dynamical system}  
\end{align}
where $\Bar{\mA}$ and $\Bar{\mB}$ are the discretized versions\footnote{Using zero-order hold and $dt=1$, exact discretization gives $\Bar{\mA} = e^{\mA}$ and  $\Bar{\mB} = \mA^{-1}(e^{\mA} - \mI)\mB$.} of $\mA$ and $\mB$. These matrices are usually frozen during training, although they need not be. The input to the LTI system, $u(t)$, is computed by projecting the input to the RNN cell, $\vx_{t} \in R^{d_x}$, the hidden state, $\vh_{t} \in \R^{d_h}$, and the memory state, $\vm_{t} \in \R^{d}$, onto their respective encoding weights $(\ve_{x}, \ve_{h}, \text{and
}\ve_{m})$. The memory state, hidden state and the input to the cell are then combined using the weight matrices $(\mW_{x},\mW_{h}, \text{and } \mW_{m})$ and passed through the nonlinearity, $f$. The encoding vectors and the kernel matrices are learned during training. 

\subsection{Our Model}\label{section_ff_LMU}

In this paper, we propose a modified LMU architecture by making two changes to the equations defined above. First, we remove certain connections in (\ref{input}) and (\ref{nonlinear dynamical system}), namely $(\ve_{h}^{T}, \ve_{m}^{T})$ and $(\mW_{h})$; this is done to aid parallelization. Second, instead of projecting the input to the RNN cell, $\vx_t$ down to a scalar as in (\ref{input}), we implement a general affine transformation followed by an element-wise nonlinearity; the original architecture is better suited for dealing with 1D or low-dimensional inputs, and this is a straightforward generalization of the encoding equation for higher dimensional inputs. Additionally, adding a bias term to equation (\ref{nonlinear dynamical system}), we end up with the following:
\begin{align}
    \displaystyle
    & \vu_{t} =  f_1(\mU_x \vx_{t} + \vb_u)\label{input_ours},\\
    &\vm_{t} = \Bar{\mA} \vm_{t-1} + \Bar{\mB} \vu_{t},\label{DN_ours}\\
    &\vo_{t} = f_2(\mW_{m} \vm_{t} + \mW_{x} \vx_{t} + \vb_o)\label{output_ours}.  
\end{align}
Note that equations (\ref{input_ours}) and (\ref{output_ours}) are equivalent to having {\it time-distributed} dense layers before and after equation (\ref{DN_ours}). In general, we expect our model to be modified and used in combination with other feed-forward layers, including self-attention. For example, we found a gated architecture \citep{srivastava2015training} where equation (\ref{input_ours}) is modified to
\begin{align*}
    \vu_{t} =  f_1(\mW_u \vx_{t} + \vb_u) \cdot \vg +   \vx_t \cdot \left(1 - \vg\right),
\end{align*}
to work well for the addition problem (results not shown). The gate is defined as $\vg = \sigma(\mW_g \vx_t + \vb_g)$, i.e., a sigmoid-activated affine transformation where the bias is initialized to -1.

Additionally, now that the input to the LTI system in this case is a vector, $\vu_t \in \R^{1\times d_u}$, we have that $\vm_t \in \R^{d \cdot d_u}$, and equation (\ref{DN}) can be thought of as implementing the following: 
\begin{align}
\vm_t = \text{reshape}(\Bar{\mA}  \text{ reshape}(\vm_{t-1},  (d, \,d_u)) + \Bar{\mB} \vu_t, \,\,d\cdot d_u).
\end{align}


\paragraph{Parallel Training} One of the motivations for the above mentioned architectural changes is that the model now has only one recurrent connection: $\vm_t$'s dependence on itself from the past in equation (\ref{DN_ours}). But because this is an LTI system, standard control theory \citep{aastrom2010feedback} gives us a non-iterative way of evaluating this equation as shown below\footnote{Restricting ourselves to 1D input for the purposes of illustration.} 
\begin{align}
    \vm_t =  \sum_{j=1}^{t} \bar{\mA}^{t-j} \bar{\mB} u_j.
\end{align}
Defining $\mH = \begin{bmatrix} \Bar{\mA}^0 \bar{\mB} & \Bar{\mA}  \bar{\mB} & \hdots \end{bmatrix} \in \mathbb{R}^{d \times n}$ and 
\begin{align}
    \mU = \begin{bmatrix} u_1 & u_2 & u_3 & \hdots & u_n \\
                              & u_1 & u_2 & \hdots & u_{n-1}\\ 
                              &     & u_1 & \hdots & u_{n-2} \\
                              &     &     & \ddots & \vdots \\
                              &     &     &         &  u_1\end{bmatrix} \in \mathbb{R}^{n \times n},
\end{align}
the above convolution equation can alternatively be written as a matrix multiplication:
\begin{align} \label{parallel_time}
    \vm_{1:n} = \mH \mU,
\end{align}
where $n$ is the sequence length. Given that $\mH$ is the impulse response of the LTI system, in practice we compute $\mH$ by feeding in an impulse to the RNN version of the DN (equation (\ref{DN_ours})). Note that in our method the $\bar{\mA}$ and $\bar{\mB}$ matrices are frozen during training, so the impulse response need only be computed once. In case of multi-dimensional inputs, we would repeat the above computation several times, once for each dimension of the input. It is also evident from the structure of the $\mU$ matrix that although this reformulation turns the DN into a feedforward layer, it still respects causality. In other words, the state $\vm_t$ depends only on the inputs seen until that point of time, i.e., $u_i : i \leq t$.   

\begin{table}
\caption{Complexity per layer and minimum number of sequential operations of various architectures. $n$ is the sequence length, $d_x$ is the input dimension, $d$ is the order, and $k$ is the size of the kernel. First three rows are as reported in \citet{vaswani2017attention}.}
\label{table:complexity}
\begin{center}
\begin{tabular}{lcc}
\hline\noalign{\smallskip}
Layer Type & Complexity & Sequential Ops \\
\hline\noalign{\smallskip}
RNN & $O(n \cdot d^2_x)$ & $\surd$ \\
Convolution & $O(k \cdot n \cdot d_x^2)$ & $\times$  \\
Attention & $O(n^2 \cdot d_x)$ & $\times$ \\
\hline\noalign{\smallskip}
DN (\ref{DN_ours}) &  $O(n \cdot d^2 \cdot d_x)$ & $\surd$\\
DN (\ref{parallel_time}) &  $O(n^2 \cdot d \cdot d_x)$ & $\times$\\
DN (\ref{parallel_time_final}) &  $O(n \cdot d \cdot d_x)$ & $\times$\\
DN (\ref{parallel_frequency}) &  $O(n \cdot \log{n} \cdot d \cdot d_x)$  & $\times$ \\
\hline
\end{tabular}
\end{center}
\end{table}

\paragraph{Complexity} With the new formulation, we immediately see that it is computationally (i.e., in terms of the number of operations) advantageous in situations where we only need the final state (\verb|return_sequences=False| in Keras terminology). Instead of using (\ref{DN_ours}) to explicitly simulate the first $n-1$ steps in-order to compute the state at the time-step $n$, we can instead just compute \begin{align}\label{parallel_time_final}
    \vm_n = \mH \mU_{:n},
\end{align}
thus reducing the complexity of computing the final state, $\vm_n$, from $O(n \cdot d^2 \cdot d_x)$ to $O(n \cdot d \cdot d_x)$, where $n$ is the sequence length, $d$ is the order, and $d_x$ is the dimension of the input.  We show in Section (\ref{section: Training time}) that using this implementation results in up to 200x speedup.

The more general computation (\ref{parallel_time}), although parallelizable, results in a complexity of $O(n^2 \cdot d \cdot d_x)$. This can be made more efficient by employing the convolution theorem which gives us an equivalent way of evaluating the convolution in the Fourier space as\footnote{Assuming a padded Fourier transform across the appropriate axis and automatic broadcasting when computing element-wise multiplication. See \texttt{LMUFFT} code for  details.}
\begin{align}\label{parallel_frequency}
    \vm_{1:n} = \mathcal{F}^{-1}\{\mathcal{F}\{\mH\} \cdot \mathcal{F} \{\mU_{:n}\}\}.
\end{align}
Thanks to the fast Fourier transform, the above operation has a complexity of $O(n \cdot \log_2{n} \cdot d \cdot d_x)$. These, other algorithms, and their complexities are reported in Table (\ref{table:complexity}).

It was argued in \citet{vaswani2017attention} that a self-attention layer is cheaper than an RNN when the representation dimension of the input, $d_x$, is much greater than the length of the sequence, $n$, which is seen in NLP applications. For example, standard word or sub-word based machine translation systems work with sequence lengths of about 100 and representation dimension ranging from 300 to 512. The same argument holds for the DN layer too, since we have found the inequality $\log_2 n \cdot d \ll d_x$ to hold in all our word-based NLP experiments -- this excludes text8, which works at the level of characters.  More specifically, we use $d \leq 4$ for our word-based models.

\paragraph{Recurrent Inference} Machine learning algorithms are usually optimized for training rather than deployment \citep{crankshaw2019design}, and because of that models need to be modified, sometimes non-trivially, to be more suitable for inference. For example, one of the metrics that is crucial for applications such as online Automatic Speech Recognition (ASR) is low latency. Transformers have shown faster training times and better performance than RNNs on ASR, but since they employ (global) self-attention, which requires the entire input to begin processing, they are not natural fits for such a task \citep{wang2020reducing}. Look-ahead \citep{zhang2020transformer} and chuck-based \citep{miao2020transformer} approaches are usually used to alter the architecture of self-attention for such purposes. While our model can be trained in parallel, because of the equivalence of equations (\ref{DN_ours}) and (\ref{parallel_frequency}), it can also be run in an iterative manner during inference, and hence can process data in an online or streaming fashion during inference.

\section{Experiments}

In the following experiments we compare our model against the LMU, LSTMs and transformers. With these experiments, we focus on benchmarking rather than establishing new state-of-the-art results. Hence, we stick to simple architectures, constrain ourselves to train all the models, with the exception of text8, using the Adam optimizer \cite{kingma2014adam} with all the default parameter settings. For text8, we found it helpful to reduce the learning rate by a factor of 10 halfway into training. Although unusual, we use the default optimization settings even when transfer learning. 

With models we compare against, we use results found in published work, even when they make use of more sophisticated architectures, learning schedules or regularization schemes.

We do not consider the capacity task from the original LMU paper here as they employ just the delay layer without nonlinear dynamics in order to deal with this problem, which makes their architecture essentially the same as ours.

\subsection{psMNIST}

 The permuted sequential MNIST (psMNIST) dataset was introduced by \citet{le2015simple} to test the ability of RNNs to model complex long term dependencies. psMNIST, as the name suggests, is constructed by permuting and then flattening the $(28 \times 28)$ MNIST images. The permutation is chosen randomly and is fixed for the duration of the task, and in order to stay consistent, we use the same permutation as \citet{chandar2019towards} and \citet{voelker2019legendre}. The resulting dataset is of the form (samples, 784, 1), which is fed into a recurrent network sequentially, one pixel at a time. We use the standard 50k/10k/10k split. 

\paragraph{Architecture} As was pointed out in \citet{voelker2019legendre}, in order to facilitate fair comparison, RNN models being tested on this task should not have access to more the $28^2=784$ internal variables. Keeping that in mind, and in order to make direct comparisons to the original LMU model, we consider a model with $d=468$ dimensions for memory. We set the dimension of the output state to $346$ and use $\theta=784$. Our model uses 165k parameters, the same as all the models reported in Table (\ref{table_psmnist_score}), except for the original LMU model, which uses 102k parameters, and the HiPPO-LegS model, which is reported to use 512 hidden dimensions (number of parameters is unknown).

\paragraph{Results \& Discussion} Test scores of various models on this dataset are reported in Table (\ref{table_psmnist_score}). Our model not only surpasses the LSTM model, but also beats the current state-of-the result of 98.3\% set by HiPPO-LegS \citep{gu2020hippo} recently. Thus, our model sets a new state-of-the art result for RNNs of 98.49\% on psMNIST. It is interesting that our model,
despite being simpler than the original LMU, outperforms it on this dataset. This suggests that the main advantage of the LMU over past models is the quality of its temporal memory, implemented by the DN. 

\begin{table}
\parbox{.45\linewidth}{
\centering
\caption{ psMNIST results. The first three rows are from  \citet{voelker2019legendre}, and the fourth row is from \citet{gu2020hippo}.}
\label{table_psmnist_score}
\begin{tabular}{lc}
\hline\noalign{\smallskip}
Model & Accuracy\\
\hline\noalign{\smallskip}
LSTM & 89.86\\
NRU & 95.38\\
LMU & 97.15\\
HiPPO-LegS & 98.3\\
Our Model &  {\bf 98.49}  \\
\hline
\end{tabular}

}
\hfill
\parbox{.45\linewidth}{
\centering
\caption{Mackey-Glass results.}
\label{table: mackey-glass}
\begin{tabular}{lc}
\hline\noalign{\smallskip}
Model & NRMSE  \\
\hline\noalign{\smallskip}
LSTM        &  0.059       \\
LMU         &  0.049      \\
Hybrid      &  0.045        \\
Our Model   &  {\bf 0.044} \\
\hline
\end{tabular}

}
\end{table}

\subsection{Mackey-Glass}
Mackey-Glass equations are a set of two nonlinear differential equations that were originally developed to model the quantity of mature cells in the blood. The second of these equations is interesting because it can result in chaotic attractors and is used to construct the dataset at hand. This is a time-series prediction task where we need to predict 15 time-steps into the future.

\paragraph{Architecture} \citet{voelker2018improving} compare three architectures on this dataset. The first one uses 4 layers of LSTMs $(h=28)$, the second one replaces LSTMs with LMUs $(d=4, \theta=4)$, and the final one replaces the first and third layers in the first model with LMUs. We use a relatively simple architecture where we combine our model (single layer) with an additional dense layer. We use $d=40$, $\theta=50$, and 1 and 140 units in the input and output layers, with the additional dense layer containing 80 units. We did not try other variations. All the models contain about 18k parameters and are run for 500 epochs.

\paragraph{Results \& Discussion} NRMSE scores on the test set are presented in Table (\ref{table: mackey-glass}). We see that our model outperforms the other three models in accuracy and training time. In the original paper \citet{voelker2018improving} hypothesize that the superior performance of the hybrid model is due to the presence of gates, but given that our model lacks gating mechanisms, we think that it might have to do with the LMU model being misparametrized. 

\subsection{Sentiment, Semantics and Inference} 

In this section, we explore the supervised and semi-supervised capabilities of our model. More specifically, we first look at the tasks of sentiment analysis (IMDB), semantic similarity (QQP) and natural language inference (SNLI), and then improve upon the IMDB score using a language model pre-trained on the (unlabelled) Amazon Reviews dataset \cite{ni2019justifying}.

\subsubsection*{Supervised}

\paragraph{IMDB} The IMDB dataset \citep{maas2011learning} is a standard sentiment classification task containing a collection of 50k highly polar reviews, with the training and testing sets containing 25k reviews each. We use the standard pre-processed dataset available from the Keras website,\footnote{https://keras.io/api/datasets/imdb/.} consider a vocabulary of 20k words, and set the maximum sequence length to 500 words.

\paragraph{QQP} For Quora Question Pairs (QQP), given a pair of sentences, the task is to identify whether the two sentences are semantically similar. In this case, we experiment on two train/dev/test splits: 390k/8k/8k like in \citet{shen2018baseline},  and 280k/80k/40k like in \citet{sharma2019natural}. We use a vocabulary of 20k words and truncate the sentences to be less than 25 words.

\paragraph{SNLI} The Stanford Natural Language Inference\footnote{ Dataset and published results are available at https://nlp.stanford.edu/projects/snli/.} was released to serve as a benchmark for evaluating machine learning systems on the task of natural language inference. Given a premise, the task is to determine whether a hypothesis is true (entailment), false (contradiction), or independent (neutral). We use the standard 550k/10k/10k split, consider a vocabulary of 20k words, and set the maximum sequence length to 25 words. 

\paragraph{Architecture} In our experiments, confusingly, we found the use of the DN, without any nonlinearities, to work well. Therefore, we construct parameter efficient models that employ just the DN layer, with $d = 1$ and $\theta = \texttt{maxlen}$. We use 300D Glove embeddings (840B Common Crawl; \citet{pennington2014glove}) for all our models. For the IMDB task, which is a single sentence task, we encode the sentence and pass the encoded vector to the final classification layer. For two-sentence tasks, QQP and SNLI, we encode the two sentences to produce two vectors, and then pass the vector obtained by concatenating the two vectors, their absolute difference, and their element-wise product to the final classification layer. We compare our models against the LSTM models described in \cite{gu2020hippo} for IMDB, both \cite{shen2018baseline} and \cite{sharma2019natural} for QQP, and \cite{bowman2015large} for SNLI. They all use at least an order of magnitude more parameters than our models.

\paragraph{Results \& Discussion} We present the results from the three experiments in Table (\ref{table: imdb, qqp, snli}). As we can see, our simple models based on the DN alone do indeed outperform the LSTM models. It is also noteworthy that our models use significantly fewer parameters than the LSTM models: 160x on IMDB, 650x on QQP and 60x on SNLI. 

\begin{table}
\caption{IMDB, QQP and SNLI results. IMDB result is from \cite{gu2020hippo}, QQP results are from \cite{shen2018baseline} and \cite{sharma2019natural} respectively, and SNLI is from \cite{bowman2015large}.}
\label{table: imdb, qqp, snli}
\begin{center}
\begin{tabular}{lcccc}
\hline\noalign{\smallskip}
& \multicolumn{2}{c}{LSTM} &  \multicolumn{2}{c}{Our Model}\\
\hline\noalign{\smallskip}
& Param. & Acc. & Param. & Acc. \\
\hline\noalign{\smallskip}
IMDB &  50k & 87.29  &  301 & {\bf 89.10} \\
QQP & -/800k & 82.58/81.4  & 1201 &  {\bf 86.95/85.36}\\
SNLI & 220k & 77.6 & 3.6k & {\bf 78.85}  \\
\hline
\end{tabular}
\end{center}
\end{table}

\subsubsection*{Semi-Supervised}

\paragraph{Amazon Reviews} NLP over the past couple of years has been dominated by transfer learning approaches, where language models pre-trained on large corpora are used to initialize general purpose fine-tuning architectures to help with various downstream tasks. We therefore consider using a pre-trained language model on the Amazon Reviews dataset to improve the previously achieved score on IMDB. Here, we make direct comparisons to the LSTM model described in \citet{radford2017learning}. While they use the entire dataset (82 million reviews) for pre-training and train their model for a month on 4 GPUs, due to resource constraints, we use less than 5\% (3.7 million reviews) and train our model for about 12 hours on a single GPU. We use a vocabulary of 30k words.

\paragraph{Architecture} We have found the repeating block architecture, where each block is composed of our model, a few highway layers, and a dense layer, when used with skip connections to work well (see Figure in supplementary materials). Since the inputs, $d_x$, are high dimensional, using a large $\theta$, which would have to be accompanied by a large order $d$ to maintain resolution, would result in  vectors that are $d_x \cdot d$ dimensional, which is not desirable. Therefore, we instead work with smaller $\theta$ (and hence smaller $d$) and use several repeating blocks to take long-term dependencies into account. This is similar to how convolutional networks are used: small kernel sizes but many convolutional layers. If $\theta_i$ is the setting we use with the DN in the i\textsuperscript{th} block, then the effective delay, $\theta_e = \sum_i \theta_i$. In this specific case, we have $\theta_{i} = 6 \,\, \forall \,\, i$, and $\theta_e = 30$. Thus, the model has access to the past 30 tokens when making a prediction. In terms of the fine-tuning performance, like \citet{Peters:2018}, we found that using deep representations, i.e., a linear combination of representations from all the blocks, to be better than using just the output of the top block. For fine-tuning, we compute a weighted sum of the outputs from the language model, and use that to classify a given review as expressing a positive or negative sentiment. Even during fine-tuning, we use the the Adam optimizer with all the default settings. We did not experiment with smaller learning rates or other learning schedules. 

\paragraph{Results \& Discussion} Results for this experiment are presented in Table (\ref{table: Amazon, IMDB}). Despite using a much smaller pre-training dataset, training for just 12 hours, and using less than 50\% of the parameters, our model outperforms the LSTM model described in \citet{radford2017learning}. We also include a self-attention based model, DistilBert \cite{sanh2019distilbert}, for comparison; it must be noted however that DistilBert was trained on a more general yet much larger dataset (concatenation of English WikiPedia and Toronto Book Corpus).

\subsection{Language Modelling}

\begin{table}
\caption{IMDB results with pre-training. First row is from \citet{radford2017learning}, and the second row is from \citet{sanh2019distilbert}.}
\label{table: Amazon, IMDB}
\begin{center}
\begin{tabular}{lcc}
\hline\noalign{\smallskip}
Model & \# parameters (Millions) & Accuracy \\
\hline\noalign{\smallskip}
LSTM        &  75         &  92.88         \\
DistilBERT  &  66         & 92.82          \\
Our Model   &  34         & {\bf 93.20}    \\
\hline
\end{tabular}
\end{center}
\end{table}

\paragraph{text 8} We evaluate our model on character level language modelling using the text8 dataset\footnote{http://mattmahoney.net/dc/textdata.}. It contains 100MB of clean text from Wikipedia and has an alphabet size of 27. As is standard, we use the first 90MB as the training set, the next 5MB as the validation set and the final 5MB as the test set. Following \citet{mikolov2012subword} and \citet{zhang2016architectural}, we set the sequence length to 180.

\paragraph{Architecture} This architecture is similar in essence to the language model used in the above section, except for the use of deep representations; we simply work with the output from the top block.\footnote{We did not test the use of deep representations in this context.} For text8, minor changes were made to adapt the model to deal with longer sequences, which is a consequence of modelling at the level of characters, and to parameter match it to the LSTM model in \citet{zhang2016architectural}, which uses around 3.2 million weights. We employ three blocks in this case and use all three DNs with the setting $\theta = 15$. 

\paragraph{Results \& Discussion} We report the scores in bits per character in Table (\ref{table: text8, iwslt15}). We see that our model performs better than the LSTM model of similar size. A couple of observations regarding model optimization: 1) we noticed that the training loss plateaus around the twelfth epoch, so we found it helpful it to decrease the learning rate by a factor of 10 around then (this is the only dataset for which we alter the optimization parameters); 2) despite that, the training slows down after a few more epochs, and we stop the optimization after 20 epochs; we believe that a more carefully designed learning schedule might be able to help with faster convergence. We also observed that using a self-attention layer after the final block helps with generalization; this perhaps points to the fact that the model needs a context that is longer 45 tokens in order to make predictions.  

\subsection{Translation}

\paragraph{IWSLT'15 En-Vi} IWSLT'15 English to Vietnamese is a medium-resource language translation task containing 133k sentence pairs. Following \citet{luong2015stanford}, we do not perform any pre-processing other than replacing words that occur less frequently than 5 by the \texttt{<unk>} token, which leaves us with vocabulary sizes of 17k and 7.7k for English and Vietnamese, respectively. We use the TED tst2012 as the validation set and TED tst2013 as the test set. We use 300D representation for source and target embeddings.

\paragraph{Architecture} For translation, we use a standard encoder-decoder architecture inspired by the Amazon reviews language model, and we also employ an attention layer to help with translation. Our model's about the same size as the the 24 million LSTM model described in \citet{luong2015stanford}. Due to time constraints, the architecture and hyperparamters for this problem were relatively underexplored.

\paragraph{Results \& Discussion} Case sensitive BLEU scores are reported in Table (\ref{table: text8, iwslt15}). Our model brings in an improvement of 2.3 BLEU over the LSTM model. When tested on lower-cased text (without altering the training procedure), we obtained a higher score of 26.2 BLEU. One major limiting factor of this analysis is that we use a small, fixed vocabulary (17k and 7.7k words), with no way of dealing with out-of-vocabulary words. For future work, we intend to experiment with an open-vocabulary encoding scheme such as byte pair encoding \cite{sennrich2015neural}. 

\begin{table}
\caption{Language modelling and translation results. text8 score is from \citet{zhang2016architectural}, and IWSLT score is from  \citet{luong2015stanford}. ${}^a$(case sensitive), ${}^b$(case insensitive).} 
\label{table: text8, iwslt15}
\begin{center}
\begin{tabular}{lcc}
\hline\noalign{\smallskip}
Model & text8 & IWSLT'15 \\
\hline\noalign{\smallskip}
LSTM        &  1.65        &  23.3         \\
Our Model   & {\bf 1.61}    & {\bf 25.5$^a$/26.2$^b$}    \\
\hline
\end{tabular}
\end{center}
\end{table}

\subsection{Training Time}\label{section: Training time}
Here we explore the effect of parallelization on training time. We refer to architectures that implement the DN using equation (\ref{DN_ours}) as the \textit{LTI version} and the ones that use either (\ref{parallel_time_final}) or (\ref{parallel_frequency}), depending on whether \verb|return_sequences| is false or true, as the \textit{parallel version}. 

\begin{figure}
    \centering
    \includegraphics[scale=0.4]{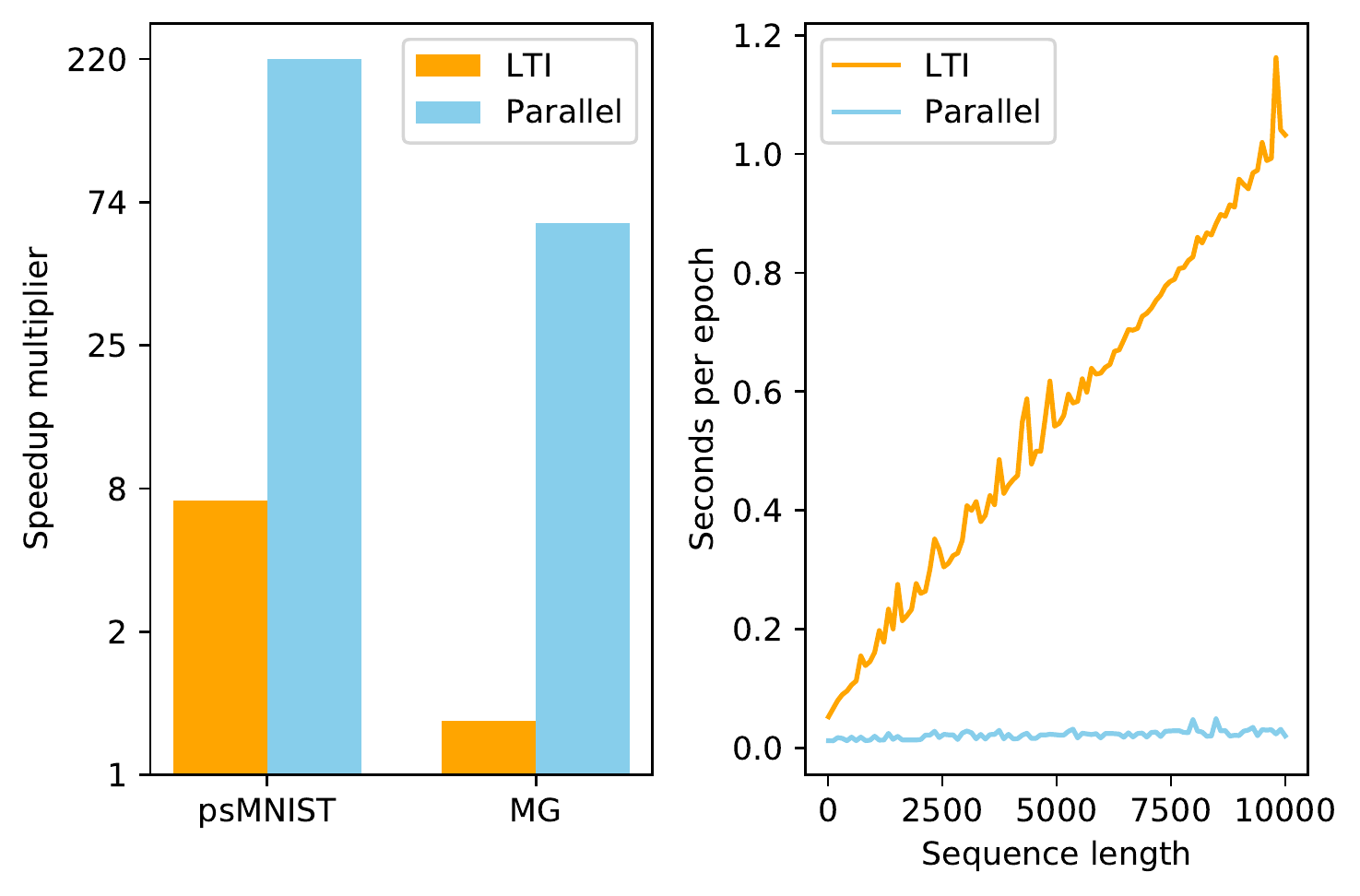}
    \caption{(left) The plot shows the speedup we obtain from switching from the LMU to the LTI (in orange) and parallel implementation (in blue) of our model. (right)  This show the effect of increasing the sequence length on the LTI and parallel versions of our model. All results were measured using a single GTX 1080. }
    \label{fig:bar}
\end{figure}

Results are presented in Figure (\ref{fig:bar}). On the left, we compare the speedup we obtain by going from the original LMU to our model, in both LTI and parallel forms, on psMNIST and Mackey-Glass. We first notice that switching from the LMU to the LTI version results in non-trivial gains. This is solely due to the reduction in the number of recurrent connections. On top of this, switching to the parallel version, owing to long sequences (784 and 5000) and, in the case of psMNIST, a reduction in the computational burden, results in substantial improvements in training times of 220x and 64x respectively.\footnote{We note that for Mackey-Glass we use a (parameter matched) one layer LMU model instead of the 4 layer model used in \citet{voelker2019legendre}; the difference with respect to the original architecture is more drastic, with our model (parallel) being about 200x faster.}

On the right of Figure (\ref{fig:bar}), we look at how varying the sequence length of the inputs effects the time it takes to complete one epoch. We see that the LTI version exhibits a linear growth whereas the parallel one essentially stays constant.  

\section{Conclusion}
With the intention of alleviating the long standing problem of speeding up RNN training, we introduce a variant of the LMU architecture that can process data in a feedforward or sequential fashion. When implemented as a feedforward layer, it can utilize parallel processing hardware such as GPUs and thus is suitable for large scale applications, and when run as an RNN, it is useful in applications where the amount of available memory is a limitation. We demonstrate the effectiveness of this architecture by experimenting on a range of tasks, of varying scale and difficulty. We also briefly consider the question of computational complexity of our model, and argue for its suitability to large scale applications in the domain of NLP, a direction we will pursue in the future. 

We note that our method of parallelization applies to all deep architectures with linear recurrent dependencies, and although we focus on a specific LTI system throughout, we hope that our analysis highlights the utility of linear systems for the purposes of machine learning. In sum, we believe that linear systems offer a scalable solution to many time series problems without sacrificing accuracy. 

\section*{Acknowledgements}
We would like to thank Aaron Voelker, Vincenzo Heska, Alison Shi, Andreas Stockel, and Terry Stewart for discussions that helped improve this paper. This work was supported by CFI and OIT infrastructure funding as well as the Canada Research Chairs program, NSERC Discovery grant 261453, AFOSR grant FA9550-17-1-0026 and OGS graduate funding.


\bibliography{example_paper}
\bibliographystyle{icml2021}

\clearpage

\section*{Supplementary Materials}
\begin{figure}[!htb]
    \centering
    \includegraphics[scale=0.25]{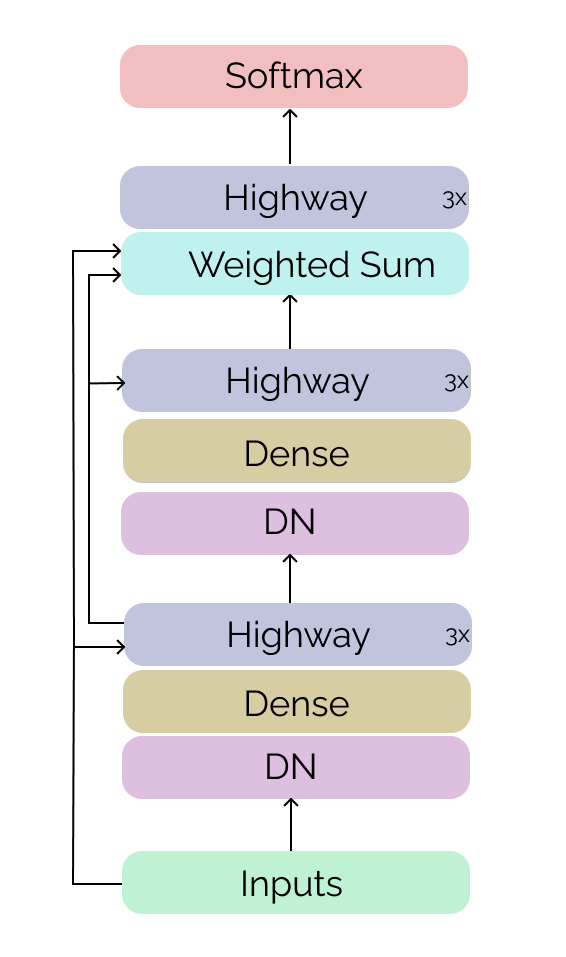}
    \caption{Illustration of the language model used for pre-training on the Amazon Reviews dataset. Although the actual model uses five blocks (combination of DN, Dense and Highway), we only show two blocks in the above figure.}
    \label{fig:amazon_pre-training}
\end{figure}

\end{document}